\documentclass{article}

\usepackage[preprint]{neurips_2024}

\usepackage[utf8]{inputenc} 
\usepackage[T1]{fontenc}    
\usepackage{hyperref}       
\usepackage{url}            
\usepackage{booktabs}       
\usepackage{amsfonts}       
\usepackage{nicefrac}       
\usepackage{microtype}      
\usepackage{xcolor}         

\usepackage{graphicx}

\usepackage{natbib}
\usepackage{amsmath}
\usepackage{multicol}
\usepackage{algorithm}
\usepackage{algpseudocode}
\usepackage{booktabs}

\title{Inverting Black-Box Face Recognition Systems via Zero-Order Optimization in Eigenface Space}

\author{
Anton Razzhigaev$^{1}$ \quad
Matvey Mikhalchuk$^{1}$ \quad
Klim Kireev$^2$ \\
\textbf{Igor Udovichenko}$^3$ \quad
\textbf{Andrey Kuznetsov}$^{1}$ \quad
\textbf{Aleksandr Petiushko}$^{3,4}$ \\
$^1$AIRI \quad $^2$Skoltech \quad $^3$MSU \quad $^4$Elea AI\\
\texttt{razzhigaev@skol.tech}
}

\definecolor{ForestGreen}{RGB}{36,135,33}
\definecolor{GothYellow}{RGB}{235,195,18}
\definecolor{GothRed}{RGB}{192,1,1}

\newcommand{\good}[1]{\textcolor{ForestGreen}{#1}}
\newcommand{\neutral}[1]{\textcolor{GothYellow}{#1}}
\newcommand{\bad}[1]{\textcolor{GothRed}{#1}}

\begin{document}

\maketitle

\begin{abstract}
  Reconstructing facial images from black-box recognition models poses a significant privacy threat. While many methods require access to embeddings, we address the more challenging scenario of model inversion using \textit{only} similarity scores. This paper introduces DarkerBB, a novel approach that reconstructs color faces by performing zero-order optimization within a PCA-derived eigenface space. Despite this highly limited information, experiments on LFW, AgeDB-30, and CFP-FP benchmarks demonstrate that DarkerBB achieves state-of-the-art verification accuracies in the similarity-only setting, with competitive query efficiency.
\end{abstract}

\section{Introduction}
Face recognition systems, while widely deployed for applications ranging from security to user convenience, expose significant privacy risks. A critical vulnerability is model inversion, where an attacker aims to reconstruct recognizable facial images of individuals known to the system (but may be unknown to the attacker). While many inversion techniques assume access to the model's internal representations (e.g., deep embeddings), a more challenging and arguably more realistic black-box scenario involves attackers who can \textit{only} query the system for similarity scores between a candidate image and a target identity. This severely restricts the information available to an adversary.

\begin{figure}
\centering
\includegraphics[width=0.7\linewidth]{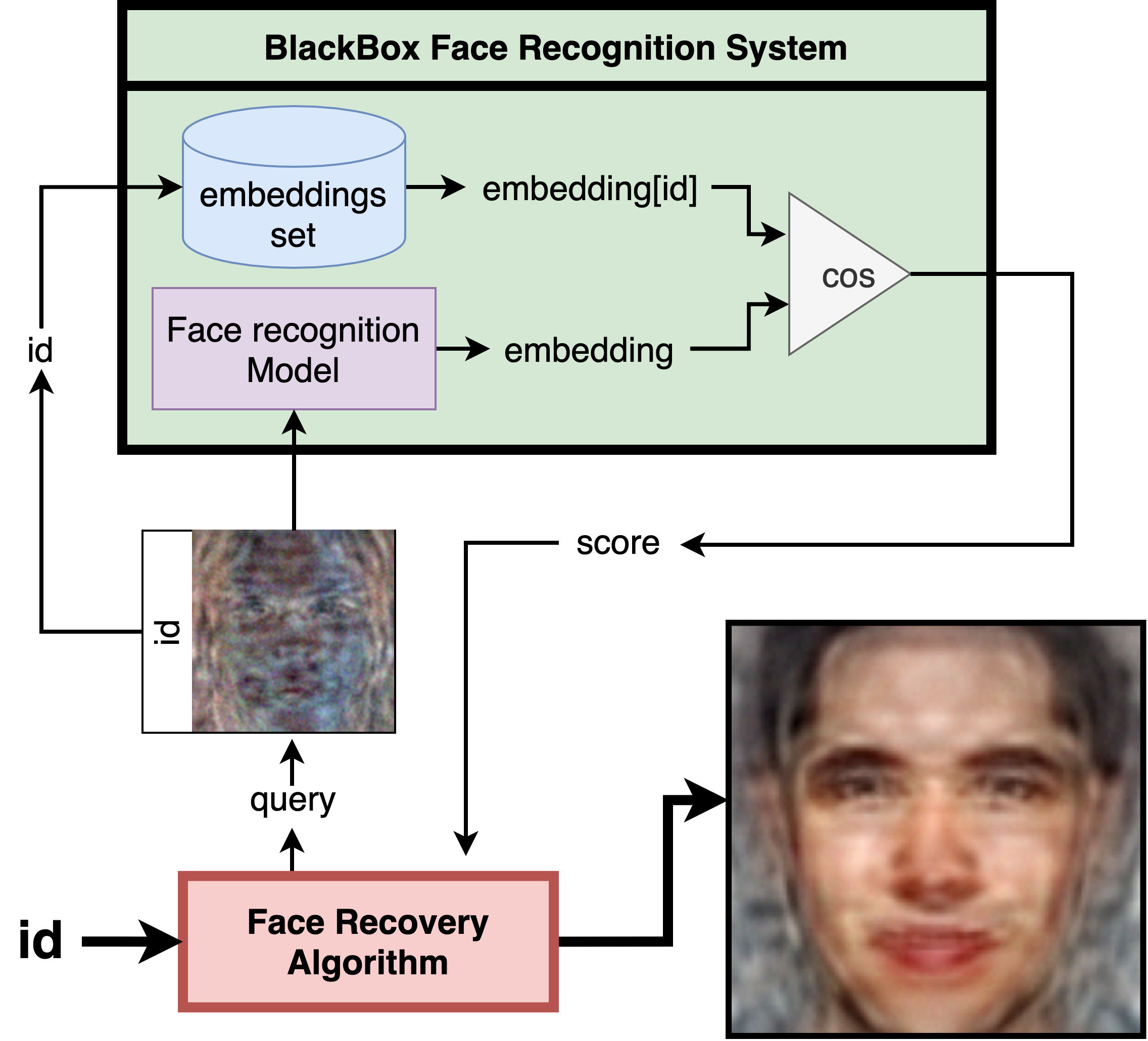}
\caption{Diagram illustrating face recovery from similarity queries to the face recognition system. By iteratively requesting similarity scores between a target person ID and a specifically generated image set, it becomes feasible to reconstruct that individual's appearance.}
\label{fig1}
\end{figure}

This paper addresses the task of face reconstruction under these stringent similarity-only black-box conditions. We introduce DarkerBB, a novel technique specifically designed for this challenging setting. DarkerBB reconstructs facial images by performing zero-order optimization within a low-dimensional latent space. This space is derived using Principal Component Analysis (PCA) applied to a public dataset of face images, effectively leveraging the "eigenfaces" concept to represent significant facial variations. Our method iteratively refines a candidate image by making queries to the target face recognition model, using only the returned similarity scores, without any knowledge of the model's architecture or direct access to its embeddings.

Figure~\ref{fig1} illustrates our face reconstruction pipeline. We premise our attack on a plausible black-box scenario: an attacker targets a specific user ID within a system, such as an online service with a profile verification feature, an API endpoint for photo similarity checks, or even a 'find look-alikes' function. The crucial element is that this system, when queried with an image generated by the attacker and the target ID, returns a similarity score reflecting how well the image matches the enrolled facial data for that ID. By iteratively crafting images and using the returned scores as feedback, the attacker refines their reconstruction, aiming to produce a recognizable facial image of the individual linked to the target ID.

\textbf{The primary contributions of our study are:}
\begin{itemize}
    \item Introduction of DarkerBB, a pioneering face reconstruction technique reliant only on similarity queries from a black-box face recognition model.
    \item Demonstration of state-of-the-art results across multiple face reconstruction benchmarks under this restrictive similarity-only setting.
    \item A method that is both query-efficient and conceptually straightforward, utilizing PCA for latent space construction and a zero-order optimization strategy.
    \item Evidence that visually recognizable faces, which also achieve high similarity scores according to independent face recognition systems, are able to be reconstructed even with such limited access.
    \item A technique to successfully reconstruct colored faces using only similarity scores unlike many other related methods operating on grayscale domain or require embedding norms.
\end{itemize}

The source code is available in GitHub\footnote{\url{https://github.com/FusionBrainLab/AdversarialFaces}}

\begin{table*}[t]
\caption{Conceptual comparison of face reconstruction methods. We emphasize the type of input required from the target model and the prior knowledge used. Our method (DarkerBB) operates in the most constrained “black-box” scenario, requiring only similarity scores.}
\label{tab:comparison_updated}
\centering
\tiny
\setlength{\tabcolsep}{4pt}
\begin{tabular}{l l l c l}
\toprule
\textbf{Method} & \textbf{Attack Type} & \textbf{Required Input from Model} & \textbf{Color} & \textbf{Prior / Search Space} \\
\midrule
\cite{mahendran2015understanding} 
& \bad{White-Box} 
& \bad{Internal states (gradients)} 
& \good{+} 
& Pixel space + regularization \\
\addlinespace
NbNet~\cite{mai2018reconstruction} 
& \good{Black-Box} 
& \neutral{Full embeddings} 
& \good{+} 
& Pixel space (CNN decoder) \\
Vec2Face~\cite{duong2020vec2face} 
& \good{Black-Box} 
& \neutral{Full embeddings} 
& \good{+} 
& GAN latent space \\
ID3PM~\cite{Kansy2023Diffusion} 
& \good{Black-Box} 
& \neutral{Full embeddings} 
& \good{+} 
& Diffusion model prior \\
DiffUMI\cite{Wang2025DiffUMI} 
& \good{Black-Box}* 
& \neutral{Full embeddings} 
& \good{+} 
& Diffusion model prior (0th-order opt.) \\
StyleGAN search~\cite{stylegansearch} 
& \good{Black-Box} 
& \neutral{Full embeddings} 
& \good{+} 
& StyleGAN latent space \\
\addlinespace
Gaussian Blobs~\cite{razzhigaev2020blackbox} 
& \good{Black-Box} 
& \neutral{Similarity + embedding norm} 
& \bad{--} 
& Linear combination of basis (Gaussians) \\
LOKT~\cite{Nguyen2023LOKT} 
& \good{Black-Box} 
& \neutral{Labels / confidence scores} 
& \good{+} 
& GAN latent space \\

\addlinespace
\textbf{DarkerBB (Ours)} 
& \good{Black-Box} 
& \good{Only similarity scores} 
& \good{+} 
& \textbf{PCA eigenfaces space} \\
\bottomrule
\multicolumn{5}{l}{\footnotesize * May be considered “gray-box” if gradient feedback is used.} \\
\end{tabular}
\end{table*}

\section{Related Works}
Deep face recognition systems have achieved remarkable performance on large-scale benchmarks, yet they remain vulnerable to model-inversion attacks~\cite{Fredrikson2015}. Early attempts demonstrated that even partial information, such as confidence scores in closed-set classification, could be exploited to reconstruct visually recognizable faces~\cite{Fredrikson2015,Galbally2010}. With the rise of embedding-based open-set face recognition, reconstruction methods have evolved to handle high-dimensional facial representations (embeddings).

A range of \textbf{training-based approaches} learns a decoder that maps embeddings back to images, assuming black-box or gray-box conditions but typically requiring a large training set. \citet{mai2018reconstruction} proposed NbNet, one of the first methods to directly invert a FaceNet embedding. More recent works leverage advanced generative models. For instance, \citet{duong2020vec2face} introduced \emph{Vec2Face} (DiBiGAN), distilling knowledge from a face-recognition network to invert embeddings while preserving identity. Similarly, \citet{Zhang2020Secret} presented a secret-revealing strategy using conditional generative adversarial networks (GANs), though initially for closed-set classification.

\textbf{GAN-based methods} are especially popular due to their ability to generate high-resolution faces. Some utilize pre-trained StyleGAN/StyleGAN2 backbones to embed the facial code into a latent space~\cite{Shahreza2023NeurIPS,Shahreza2023ICIP,Shahreza2023PAMI}, enabling photorealistic reconstruction that closely matches the target ID. Other studies refine latent codes through iterative optimization~\cite{Khosravy2022MIA,Struppek2022PPA,Qiu2024IFGMI}, but may require multiple queries or partial access to the recognition model.

Recently, \textbf{diffusion-based models} have shown strong potential. \citet{Kansy2023Diffusion} introduced ID3PM, controlling the denoising process via the target embedding. \citet{Papantoniou2024Arc2Face} proposed Arc2Face, a foundation model trained on millions of web-face images, enabling identity-preserving face generation directly from the target's ArcFace embedding. Similar ideas drive \citet{Wang2025DiffUMI}, who combine zero-order optimization and diffusion priors to invert embeddings even with limited gradient feedback. A complementary approach uses an adapter layer to align embeddings from unknown recognition models to the latent space of a universal diffusion model~\cite{Shahreza2024ArxivAdapter}. These methods highlight the versatility of current generative frameworks in performing high-fidelity face reconstruction.

\textbf{Similarity-score-based algorithms} circumvent the need for the face embedding itself, optimizing images to maximize the black-box similarity score. Although older ``hill-climbing'' methods~\cite{Galbally2010} or gradient-based strategies (in a white-box setting)~\cite{Fredrikson2015} often produce blurry results, new zero-order schemes coupled with strong priors~\cite{Wang2025DiffUMI} achieve high verification rates with fewer queries. Such label- or score-only attacks~\cite{Nguyen2023LOKT,Wu2025ABE} confirm that minimal output feedback can still reveal sensitive information. A key difference is that 'probability' or 'confidence score' attacks, exemplified by LOKT~\cite{Nguyen2023LOKT}, generally leverage the model's output from a classification task (e.g., the confidence that an image matches a specific enrolled identity). Our method, however, directly optimizes using the raw similarity score (such as cosine similarity) which quantifies feature-level resemblance between a probe and the target, a common output in open-set verification systems rather than explicit classification probabilities. Moreover, even partial leakage of embeddings (e.g., part of the feature vector) was shown to be enough for accurate face reconstruction~\cite{Shahreza2024ICASSP}.

Overall, while earlier approaches produced low-resolution, grayscale images~\cite{mai2018reconstruction,razzhigaev2020blackbox,Khosravy2022MIA}, the latest GAN- and diffusion-based methods yield photo-realistic reconstructions that reliably pass face verification protocols. These trends underscore the urgent need for more robust privacy-preserving approaches and secure embedding transformations. A brief conceptual comparison of methods is shown in Table \ref{tab:comparison_updated}.


\section{Methodology: Face Reconstruction via Zero-Order Optimization in PCA Space}

We introduce DarkerBB, a novel algorithm for reconstructing facial images from a black-box face recognition system using only similarity scores. Our approach leverages Principal Component Analysis (PCA) to define a low-dimensional latent space of faces, often referred to as "eigenfaces". The reconstruction is then framed as a zero-order optimization problem within this PCA space.

\subsection{PCA-based Latent Space}
Let a training dataset of face images be denoted by $D = \{I_1, I_2, ..., I_M\}$, where each $I_j$ is a flattened image vector. We apply PCA to this mean-centered dataset to find a set of $k$ principal components (eigenfaces) that capture the most significant variations in facial appearance. These eigenfaces form an orthogonal basis $E = [e_1, e_2, ..., e_k]$, where each $e_i \in \mathbb{R}^d$ and $d$ is the dimensionality of the flattened image. Any face image $I$ can then be approximated as a linear combination of these eigenfaces plus a mean face $\mu$:
$$I \approx \mu + \sum_{i=1}^{k} c_i e_i = \mu + E c$$
where $c = [c_1, c_2, ..., c_k]^T$ is the vector of PCA coordinates in the $k$-dimensional latent space. The eigenfaces were normalized so that the PCA coordinates $c$ computed from the training data have variance equal to 1 along each axis (i.e., $\mathrm{Var}(c) = \mathbf{I}$). For our work, we use $k=1024$, with eigenfaces derived from the FFHQ dataset \cite{karras2019style}.

\subsection{Zero-Order Optimization Algorithm}
The core of DarkerBB is an iterative optimization process to find the PCA coordinates $c$ that generate a face image $I(c)$ maximizing the similarity score $S(I(c), id_{target})$ with respect to a target identity $id_{target}$, as provided by the black-box face recognition model.

The optimization proceeds as follows:
Let $c^{(t)}$ be the PCA coordinate vector at iteration $t$.
To estimate the gradient of the similarity function $S$ with respect to $c$, we employ a 2-point gradient estimator \cite{gasnikov2023randomized}, a common technique in zero-order optimization. At each iteration, we sample a random direction vector $u \in \mathbb{R}^k$ from a spherically symmetric multivariate Gaussian distribution $u \sim \mathcal{N}(0, \sigma^2\mathbf{I}_k)$. A small perturbation scalar $\sigma > 0$ (in our experiments, $\sigma = 0.3$) is used to query the similarity function at two points along this direction:
$$s_2 = S(\mu + E (c^{(t)} - u), id_{target})$$
$$s_1 = S(\mu + E (c^{(t)} + u), id_{target}),$$

where $S(\cdot,\cdot)$ denotes the cosine similarity score returned by the black-box face recognition system between the target identity and the query image.

The gradient $g^{(t)}$ in the direction of $u$ is estimated using a central difference:
$$\hat{\nabla}_{u}S(c^{(t)}) = \frac{s_2 - s_1}{2\sigma}$$
The full gradient estimate $G(c^{(t)})$ used for the update is:
$$G(c^{(t)}) = k\hat{\nabla}_{u}S(c^{(t)}) \cdot u = k\frac{s_2 - s_1}{2\sigma} u$$

The PCA coordinates are then updated using a standard stochastic gradient ascent step with a learning rate $\eta$ targeted to maximize the similarity:
$$c^{(t+1)} = c^{(t)} + \eta G(c^{(t)})$$

The process begins with an initial coordinate vector $c^{(0)}$, typically the zero vector (representing the mean face). Each iteration requires two queries to the black-box model to evaluate $s_1$ and $s_2$. The complete method is described in Algorithm \ref{alg:darkerbb}.

\begin{algorithm}[H]
\caption{DarkerBB: Face Reconstruction Algorithm}\label{alg:darkerbb}
\textbf{Input:} Target identity $id_{target}$, black-box similarity function $S(\cdot, \cdot)$, PCA basis $E$, mean face $\mu$, learning rate $\eta$, smoothing parameter $\sigma$, number of restarts $N_{restarts}$, restart iterations $N_{restart\_iter}$, total optimization iterations $N_{main\_iter}$.
\begin{algorithmic}[1]
\State Initialize best overall coordinates $c_{best\_overall} \leftarrow \mathbf{0}$
\State Initialize best overall similarity $s_{best\_overall} \leftarrow -\infty$

\For{$r \leftarrow 1$ to $N_{restarts}$}
    \State Initialize current coordinates $c_{current} \leftarrow \mathbf{0}$ \Comment{Or other random initialization}
    \For{$i \leftarrow 1$ to $N_{restart\_iter}$} \Comment{Initial optimization phase}
        \State Sample random direction $u \sim \mathcal{N}(0, \sigma^2\mathbf{I}_k)$
        \State $s_1 \leftarrow S(\mu + E (c_{current} - u), id_{target})$
        \State $s_2 \leftarrow S(\mu + E (c_{current} + u), id_{target})$
        \State $G \leftarrow k\frac{s_2 - s_1}{2\sigma} u$
        \State $c_{current} \leftarrow c_{current} + \eta G$
    \EndFor
    \State $s_{current\_restart} \leftarrow S(\mu + E c_{current}, id_{target})$
    \If{$s_{current\_restart} > s_{best\_overall}$}
        \State $s_{best\_overall} \leftarrow s_{current\_restart}$
        \State $c_{best\_overall} \leftarrow c_{current}$
    \EndIf
\EndFor

\State Initialize final coordinates $c_{final} \leftarrow c_{best\_overall}$
\For{$i \leftarrow 1$ to $N_{main\_iter}$} \Comment{Main optimization phase}
    \State Sample random direction $u \sim \mathcal{N}(0, \sigma^2\mathbf{I}_k)$
    \State $s_1 \leftarrow S(\mu + E (c_{final} - u), id_{target})$
    \State $s_2 \leftarrow S(\mu + E (c_{final} + u), id_{target})$
    \State $G \leftarrow k\frac{s_2 - s_1}{2\sigma} u$
    \State $c_{final} \leftarrow c_{final} + \eta G$
\EndFor
\end{algorithmic}
\textbf{Output:} Reconstructed face image $I_{recon} = \mu + E c_{final}$
\end{algorithm}

\subsection{Multi-Start Strategy}
\begin{figure}[H]
\centering
\includegraphics[width=0.85\linewidth]{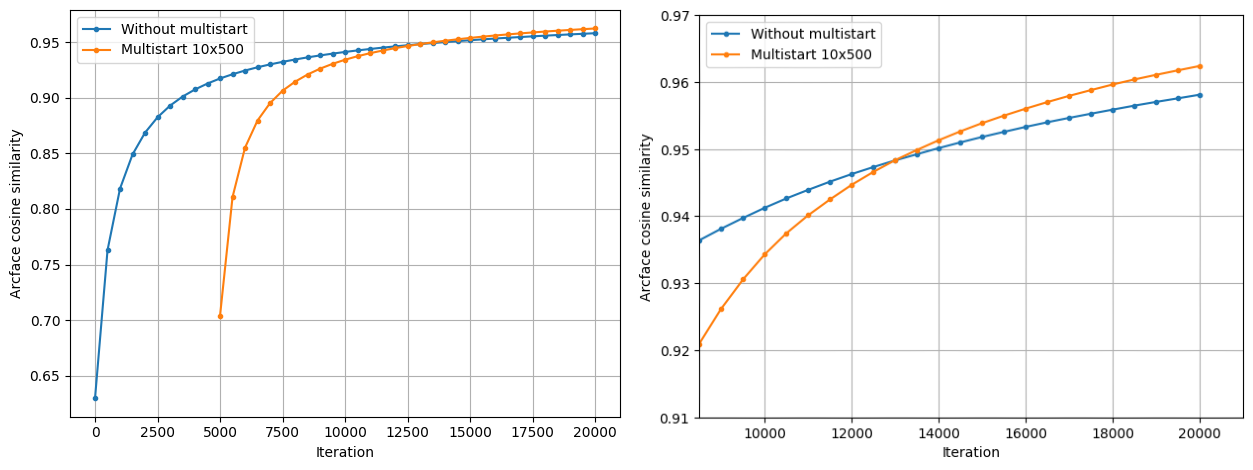}
\caption{Multi-start strategy leads to higher final similarity scores by selecting better initialization among multiple optimization paths, under the same total number of queries to the FR system.}
\label{multistart}
\end{figure}

To mitigate the risk of converging to poor local optima, we employ a multi-start strategy. The optimization is initiated $N_{restarts}$ times (e.g., 10 restarts) for a small number of iterations $N_{restart\_iter}$ each (e.g., 500 iterations). The set of PCA coordinates that yields the highest similarity score after these initial runs is chosen as the starting point for a longer, main optimization phase of $N_{main\_iter}$ iterations (e.g., 15,000). This totals approximately 20,000 optimization steps per reconstructed face, corresponding to 40,000 queries to the similarity function. This strategy is crucial, as the initial convergence trajectory often indicates the potential quality of the final reconstruction. Figure \ref{multistart} illustrates the benefit of this multi-start approach by showing how different initializations lead to varied optimization paths, with the multi-start selecting a more promising one.

The key advantage of optimizing in the PCA latent space is that it inherently restricts the search to plausible face-like images, acting as a strong prior. This allows us to effectively reconstruct faces using only similarity scores, without requiring direct access to embeddings or their norms, unlike some previous methods.

\section{Experimental Setup and Evaluation}

\subsection{Datasets and Benchmarks}
We evaluate our method on three standard face verification datasets:
\begin{itemize}
    \item \textbf{Labeled Faces in the Wild (LFW)} \cite{LFWTech}: A widely used benchmark for unconstrained face verification.
    \item \textbf{AgeDB-30} \cite{moschoglou2017agedb}: A dataset focusing on age-separated face verification.
    \item \textbf{CFP-FP} \cite{cfp-paper}: A dataset with frontal-profile face pairs.
\end{itemize}
For each dataset, experiments consist of: (1) testing on original image pairs, and (2) testing with one image in each positive pair replaced by our reconstructed version, evaluated against both ArcFace \cite{deng2019arcface} and FaceNet \cite{Schroff_2015} embeddings.

\subsection{Evaluation Protocol}
The structure of the test sets involves pairs of photos with a binary label (0 for different individuals, 1 for the same individual).
Verification accuracy is measured using K-Fold cross-validation with $k=10$. In each fold, 9 parts are used to determine the optimal similarity threshold for the best accuracy, and the remaining part is used for testing. The final accuracy is the average over the 10 folds.

When testing reconstructed images in positive pairs, the first photo of the pair is replaced by the version reconstructed by DarkerBB. The similarity score is then computed between the reconstructed image and the original second image of the pair.

\subsection{Embedding Generation and Similarity}
During verification tests, embeddings are generated by concatenating the embedding of the aligned face image with the embedding of its horizontally flipped version. The similarity between two embeddings is calculated using cosine similarity: $s(e_{1},e_{2}) = e_{1}^{T}e_{2}/||e_{1}||\cdot||e_{2}||$. This setup is consistent with practices described in related works such as \cite{Kansy2023Diffusion}.

\begin{figure}[t]
\centering
\includegraphics[width=0.8\linewidth]{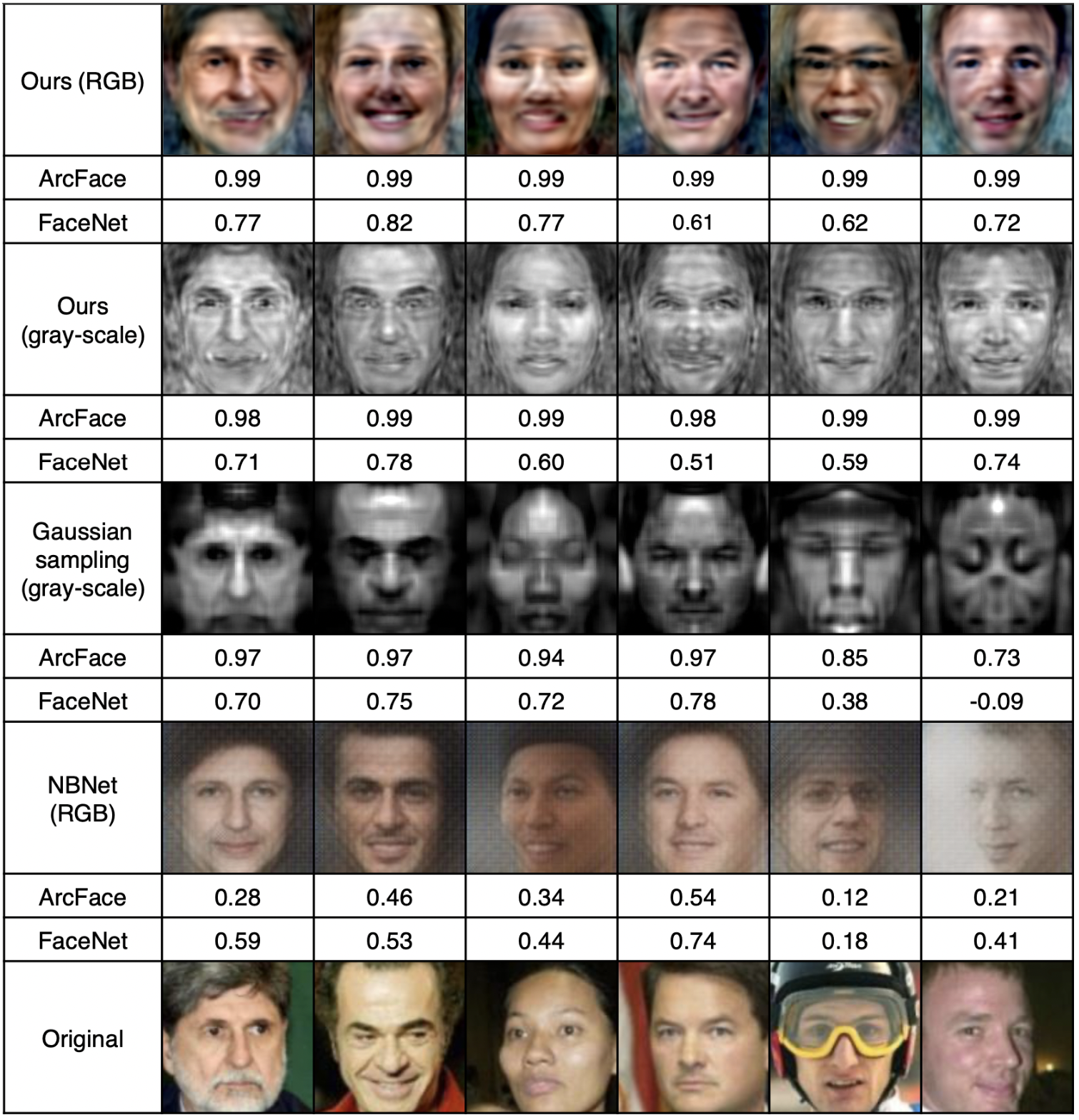}
\caption{Examples of recovered faces from LFW dataset and corresponding similarities from Arcface and FaceNet.}
\label{compare_all}
\end{figure}

\subsection{Compute Resources}

All experiments were conducted using \textbf{4× NVIDIA A100 GPUs}. We performed face reconstruction and evaluation across three benchmark datasets (LFW, AgeDB-30, CFP-FP), using two alignment methods (MTCNN, RetinaFace) and two target models (ArcFace, FaceNet), totaling 6 experimental setups.

Each image reconstruction involved 20,000 similarity queries using our zero-order optimization algorithm with multi-start initialization. All runs were parallelized across the available GPUs.

In total, the full pipeline—including preprocessing, reconstruction, and evaluation—took approximately \textbf{40 hours} to complete across all experiments. Scripts and configuration files are provided to enable reproducibility under similar compute settings.

\section{Results and Discussion}

We conducted extensive experiments to evaluate DarkerBB's performance and compare it against existing state-of-the-art methods. The primary metric is verification accuracy on standard benchmarks, indicating how well the reconstructed faces can be identified by face recognition models.

\begin{table}[H]
\centering
\tiny 
\caption{Verification accuracy (\%) on LFW, AgeDB-30, and CFP-FP benchmarks. "ArcFace $\uparrow$" and "FaceNet $\uparrow$" denote the verification model used. "DarkerBB (ArcFace target)" means ArcFace was the target black-box model for reconstruction. The values represent the accuracy achieved when one image in a pair is replaced by the reconstructed image.}
\label{tab:main_results_updated}
\setlength{\tabcolsep}{4pt} 
\begin{tabular}{lcccccc}
\toprule
\textbf{Method} & \multicolumn{2}{c}{\textbf{LFW}} & \multicolumn{2}{c}{\textbf{AgeDB-30}} & \multicolumn{2}{c}{\textbf{CFP-FP}} \\
\cmidrule(lr){2-3} \cmidrule(lr){4-5} \cmidrule(lr){6-7}
& ArcFace $\uparrow$ & FaceNet $\uparrow$ & ArcFace $\uparrow$ & FaceNet $\uparrow$ & ArcFace $\uparrow$ & FaceNet $\uparrow$ \\
\midrule
\multicolumn{7}{l}{\textit{Baselines and Prior Art \cite{Kansy2023Diffusion}}} \\
Real images (upper limit) & 99.83\% & 99.65\% & 98.23\% & 91.33\% & 98.86\% & 96.43\% \\
NbNet~\cite{mai2018reconstruction} & 87.32\% & 92.48\% & 81.83\% & 82.25\% & 87.36\% & 89.89\% \\
Gaussian Blobs~\cite{razzhigaev2020blackbox} & 89.10\% & 75.07\% & 80.43\% & 63.42\% & 61.39\% & 55.26\% \\
StyleGAN search~\cite{stylegansearch} & 82.43\% & 95.45\% & 72.70\% & 85.22\% & 80.83\% & 92.54\% \\
Vec2Face~\cite{duong2020vec2face} & 99.13\% & 98.05\% & 93.53\% & 89.80\% & 89.03\% & 87.19\% \\
ID3PM (FaceNet target)~\cite{Kansy2023Diffusion} & 97.65\% & 98.98\% & 88.22\% & 88.00\% & 94.47\% & \textbf{95.23\%} \\
ID3PM (InsightFace target)~\cite{Kansy2023Diffusion} & 99.20\% & 96.02\% & 94.53\% & 79.15\% & 96.13\% & 87.43\% \\
\midrule
\multicolumn{7}{l}{\textit{Our Experiments}} \\
Real images (Our Reproduced Setup) & 99.83\% & 99.18\% & 98.22\% & 91.55\% & 98.41\% & 95.74\% \\
\textbf{DarkerBB (ArcFace target)} & \textbf{99.78\%} & 92.23\% & \textbf{98.10\%} & 82.00\% & \textbf{98.14\%} & 83.23\% \\
\textbf{DarkerBB (FaceNet target)} & 95.80\% & \textbf{99.07\%} & 83.47\% & \textbf{90.58\%} & 89.76\% & 95.01\% \\
\bottomrule
\end{tabular}
\end{table}

The results in Table \ref{tab:main_results_updated} and some visualization comparison of publicly available methods in Figure \ref{compare_all} demonstrate that even with access limited to only similarity scores, DarkerBB can reconstruct facial images that are identifiable by state-of-the-art face recognition systems. The use of PCA provides a strong prior, guiding the optimization towards facial structures, while the zero-order method effectively navigates the search space using only the limited feedback. The multi-start strategy further enhances the robustness by mitigating convergence to suboptimal local minima.  

\section{Ablation Studies}
\label{sec:ablation_studies}

To ensure the robustness and optimal configuration of DarkerBB, we performed several ablation studies. These experiments investigate the impact of key hyperparameters and design choices on reconstruction performance. All ablation studies were conducted on the LFW dataset, using ArcFace as the target face recognition system for optimization and FaceNet for evaluating generalization, unless otherwise specified. The impact of the number of optimization iterations is discussed first due to its graphical representation. Subsequently, key findings for hyperparameter tuning related to PCA dimensionality, noise scale for gradient estimation, and the number of multi-start initializations are summarized in Table~\ref{tab:combined_ablations} and discussed.

\begin{figure}[H]
\centering
\includegraphics[width=0.95\linewidth]{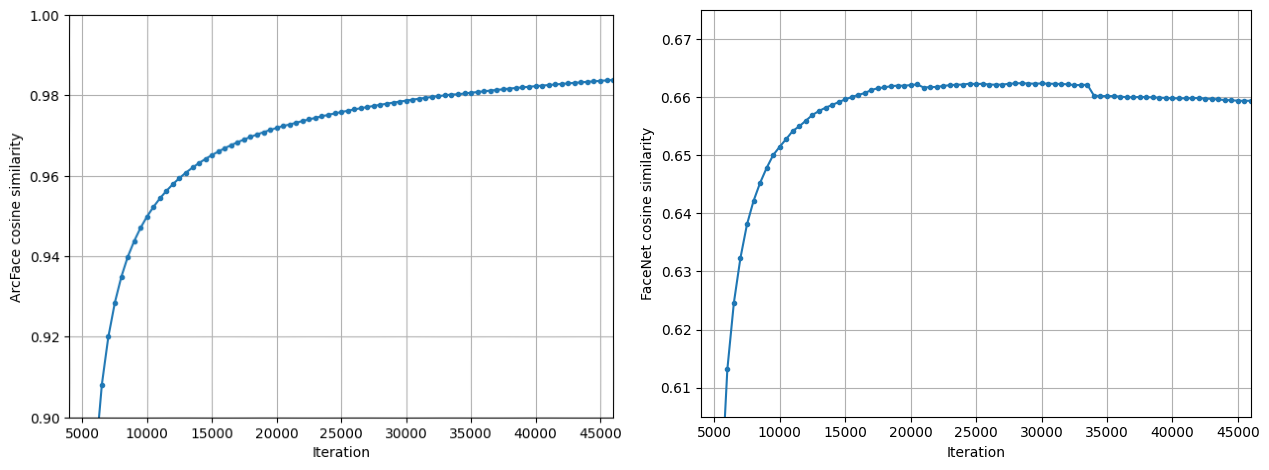}
\caption{Cosine similarity during optimization. Left: similarity measured by the target ArcFace model. Right: similarity measured by an independent FaceNet model. These results informed our choice of iteration count.}
\label{fig:iter-ablation-both-main}
\end{figure}

To determine a suitable number of optimization iterations, we analyzed the change in cosine similarity during the optimization process. Figure~\ref{fig:iter-ablation-both-main} illustrates this progression. While similarity with the target ArcFace model (left panel) continues to increase even up to 45,000 iterations, the similarity as measured by the unseen FaceNet model (right panel) tends to saturate around 20,000--25,000 iterations and may subsequently decline. This decline suggests potential overfitting to the specific characteristics of the target FR system, which could impair the generalization of the reconstructed face to other systems.

Based on this observation, and to balance reconstruction quality with generalization, we selected 20,000 optimization iterations for all main experiments reported in this paper.

\begin{table}[H]
\centering
\small
\caption{Ablation study results for key hyperparameters on LFW. For each hyperparameter, settings were varied while others were kept at their default or previously optimized values. Accuracies (\%) are reported for reconstructions optimized against ArcFace, then evaluated by ArcFace (target model) and FaceNet (transfer model). Bold values in the FaceNet column indicate the setting chosen for optimal generalization based on these ablations.}
\label{tab:combined_ablations}
\begin{tabular}{llcc}
\toprule
\textbf{Hyperparameter} & \textbf{Setting} & \textbf{ArcFace ($\uparrow$)} & \textbf{FaceNet ($\uparrow$)} \\
\midrule
PCA Dimensionality ($k$) & 256 & 99.67 & 89.52 \\
                         & 512 & 99.78 & 91.90 \\
                         & 1024 & 99.78 & \textbf{92.23} \\
                         & 2048 & 99.83 & 92.22 \\
\midrule
Noise Scale ($\sigma$)   & 0.15 & 99.83 & 91.42 \\
                         & 0.30 & 99.78 & \textbf{92.23} \\
                         & 0.60 & 99.78 & 89.43 \\
\midrule
\# Restarts ($N_{\text{restarts}}$) & 3 & 99.83 & 88.18 \\
                                 & 10 & 99.78 & \textbf{92.23} \\
                                 & 15 & 99.83 & 91.62 \\
\bottomrule
\end{tabular}
\end{table}

The dimensionality $k$ of the PCA eigenface space is a crucial parameter, dictating the expressive capacity of our facial prior. We explored its impact by testing $k \in \{256, 512, 1024, 2048\}$. The results, presented in Table~\ref{tab:combined_ablations}, show that $k=1024$ achieves an ArcFace accuracy of 99.78\% and the highest FaceNet generalization score of 92.23\% among values up to 1024 (with $k=2048$ being marginally lower at 92.22\% for FaceNet but slightly higher for ArcFace).
A qualitative assessment of visual reconstruction fidelity for different $k$ values, which also informed our choice, is provided in the Appendix (see Figure~\ref{fig:pca_k_visual} for illustrative examples).
Considering both verification performance (Table~\ref{tab:combined_ablations}) and visual quality (Appendix, Figure~\ref{fig:pca_k_visual}), $k=1024$ was selected as it offers a compelling balance.

The noise scale $\sigma$ in our zero-order optimization determines the perturbation magnitude for gradient estimation. We evaluated $\sigma \in \{0.15, 0.30, 0.60\}$. As detailed in Table~\ref{tab:combined_ablations}, $\sigma = 0.30$ provides the best generalization performance to FaceNet (92.23\%) while maintaining high accuracy (99.78\%) with the target ArcFace model. Thus, $\sigma = 0.30$ was adopted for all experiments.

DarkerBB's multi-start strategy, involving $N_{\text{restarts}}$ initial optimization runs, aims to avoid poor local optima. We tested $N_{\text{restarts}} \in \{3, 10, 15\}$. The findings in Table~\ref{tab:combined_ablations} indicate that using $N_{\text{restarts}} = 10$ significantly improves FaceNet generalization (92.23\%) over $N_{\text{restarts}} = 3$ (88.18\%). A further increase to 15 restarts did not yield additional benefits for generalization and resulted in the same ArcFace performance. Consequently, $N_{\text{restarts}} = 10$ was chosen for our standard configuration.

\section{Conclusion}
We have presented DarkerBB, a novel face reconstruction method that operates in a highly restrictive black-box scenario, requiring only similarity scores from the target face recognition system. By performing zero-order optimization in a PCA-derived eigenface space, DarkerBB successfully reconstructs color facial images that are highly recognizable by both the target model and independent verification systems. Our experiments on standard benchmarks (LFW, AgeDB-30, CFP-FP) demonstrate that DarkerBB achieves state-of-the-art verification accuracies under these stringent conditions, often comparable to reconstructions from methods requiring full embedding access.

\section{Limitations}
DarkerBB, while effective in the similarity-only black-box setting, has several limitations. The reconstruction fidelity is inherently constrained by the expressive capacity of the PCA eigenface space; faces with features underrepresented in the FFHQ dataset or very fine details might not be perfectly captured.
The requirement of approximately 40,000 queries per image can be impractical against systems with strong security measures (e.g., rate limiting) and presumes access to fine-grained similarity scores, not merely binary or coarse feedback. Finally, the zero-order optimization may be susceptible to local optima despite a multi-start strategy, and the resulting images, optimized for machine verification scores, might not always achieve perfect perceptual fidelity as judged by humans.
\vfill
\pagebreak
\appendix

\section{Appendix / supplemental material}

\begin{figure}[H]
\centering
\includegraphics[width=0.8\linewidth]{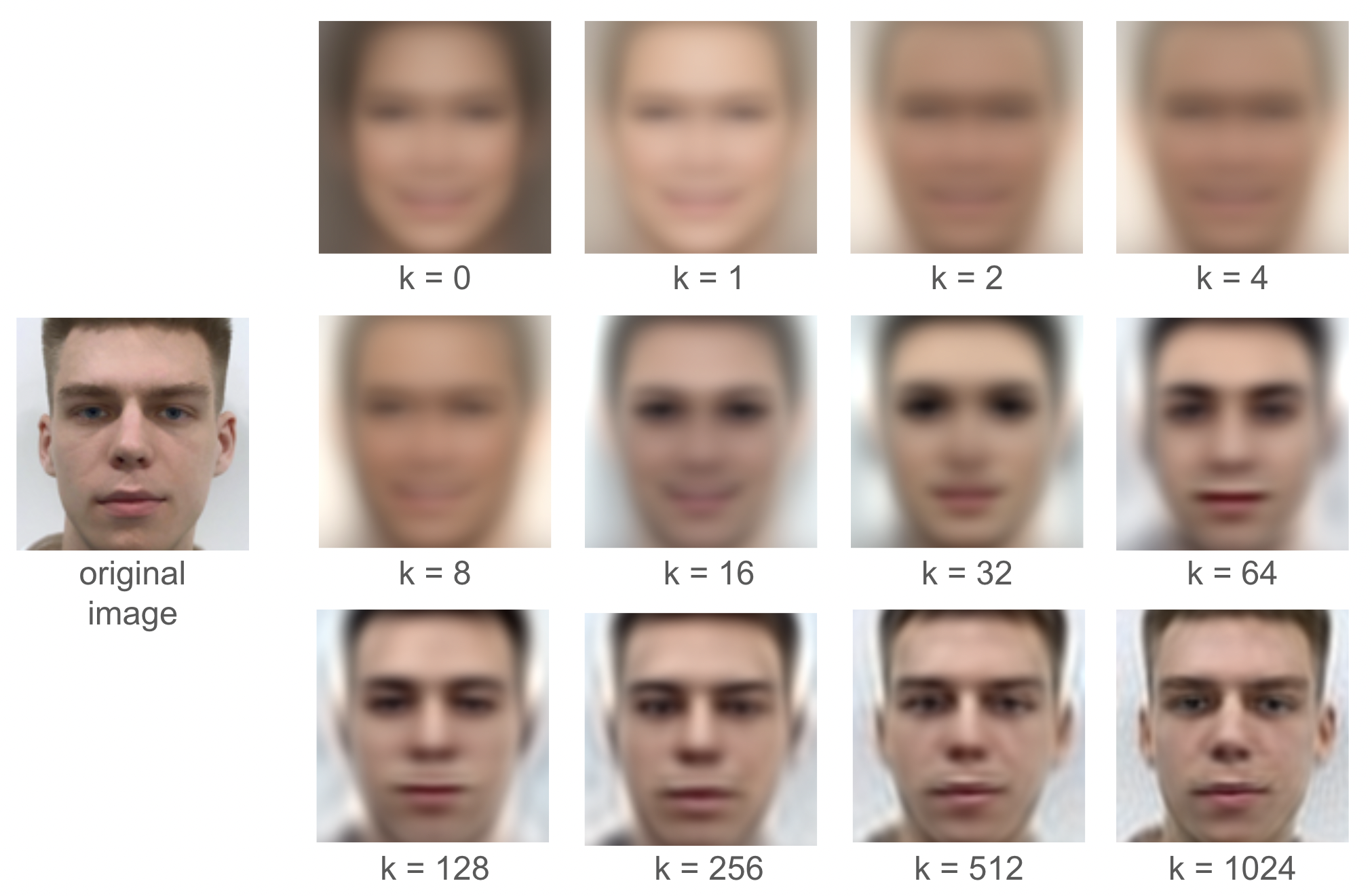}
\caption{Reconstruction of a real image using varying numbers of PCA components $k$. As $k$ increases, the reconstructed image better preserves the identity and fine details.}
\label{fig:pca_k_visual}
\end{figure}

\bibliography{biblio}
\end{document}